\documentclass[nohyperref]{article}

\usepackage{microtype}
\usepackage{graphicx}
\usepackage{subfigure}
\usepackage{booktabs} 

\usepackage{hyperref}
\usepackage{pifont}
\usepackage{colortbl}
\usepackage[dvipsnames]{xcolor}
\definecolor{Tianlong_color}{rgb}{0.858, 0.188, 0.478}

\usepackage[accepted]{icml2022}

\usepackage{amsmath}
\usepackage{amssymb}
\usepackage{mathtools}
\usepackage{amsthm}
\usepackage{makecell}
\usepackage{diagbox}
\usepackage{subfigure}
\usepackage{multirow}
\usepackage{varwidth}
\usepackage[flushleft]{threeparttable}

\usepackage{color}

\usepackage[capitalize,noabbrev]{cleveref}

\theoremstyle{plain}

\theoremstyle{definition}

\theoremstyle{remark}

\usepackage[textsize=tiny]{todonotes}

\icmltitlerunning{Linearity Grafting: Relaxed Neuron Pruning Helps Certifiable Robustness}

\begin{document}

\twocolumn[
\icmltitle{Linearity Grafting: Relaxed Neuron Pruning Helps Certifiable Robustness}



\icmlsetsymbol{equal}{*}

\begin{icmlauthorlist}
\icmlauthor{Tianlong Chen}{equal,ut}
\icmlauthor{Huan Zhang}{equal,cmu}
\icmlauthor{Zhenyu Zhang}{ut}
\icmlauthor{Shiyu Chang}{ucsb}\\
\icmlauthor{Sijia Liu}{msu,ibm}
\icmlauthor{Pin-Yu Chen}{ibm,ibm_r}
\icmlauthor{Zhangyang Wang}{ut}
\end{icmlauthorlist}

\icmlaffiliation{ut}{University of Texas at Austin}
\icmlaffiliation{cmu}{Carnegie Mellon University}
\icmlaffiliation{ucsb}{University of California, Santa Barbara}
\icmlaffiliation{msu}{Michigan State University}
\icmlaffiliation{ibm}{MIT-IBM Watson AI Lab}
\icmlaffiliation{ibm_r}{IBM Research}
\icmlcorrespondingauthor{Zhangyang Wang}{atlaswang@utexas.edu}

\icmlkeywords{Machine Learning, ICML}

\vskip 0.3in
]



\printAffiliationsAndNotice{}  

\begin{abstract}
\vspace{-0.5em}
Certifiable robustness is a highly desirable property for adopting deep neural networks (DNNs) in safety-critical scenarios, but often demands tedious computations to establish. The main hurdle lies in the massive amount of non-linearity in large DNNs. To trade off the DNN expressiveness (which calls for more non-linearity) and robustness certification scalability (which prefers more linearity), we propose a novel solution to strategically manipulate neurons, by \textit{``grafting''} appropriate levels of linearity. The core of our proposal is to first linearize insignificant ReLU neurons, to eliminate the non-linear components that are both redundant for DNN performance and harmful to its certification. We then optimize the associated slopes and intercepts of the replaced linear activations for restoring model performance while maintaining certifiability. 
Hence, typical neuron pruning could be viewed as a \textit{special case} of grafting a linear function of the fixed zero slopes and intercept, that might overly restrict the network flexibility and sacrifice its performance. Extensive experiments on multiple datasets and network backbones show that our \textbf{\textit{linearity grafting}} can ($1$) effectively tighten certified bounds; ($2$) achieve competitive certifiable robustness \emph{without certified robust training} (i.e., over $30\%$ improvements on CIFAR-10 models); and ($3$) scale up complete verification to large adversarially trained models with $17$M parameters. Codes are available at {\small\url{https://github.com/VITA-Group/Linearity-Grafting}}.

\end{abstract}


\section{Introduction}

\label{sect:intro}
Despite the prevailing successes of deep neural networks (DNNs), they remain vulnerable to adversarial examples~\citep{szegedy2013intriguing}. Therefore, certifying whether a DNN is provably robust under all bounded input perturbations are crucial for adopting DNNs in high-stake and safety-critical applications. To obtain non-trivial certified robust accuracy, 
certified adversarial defense~\citep{raghunathan2018certified,wong2018provable,mirman2018differentiable,gowal2018effectiveness,zhang2019towardsstable} which minimize a certifiable loss during training, and robustness verification methods~\citep{gehr2018ai2,dvijotham2018dual,zhang2018efficient,wang2021beta} which compute tight certified bounds for post-training networks serve as the most effective two approaches.


However, several pain points persist for these two mechanisms: ($i$) As DNNs grow larger (e.g., dozens of layers), (complete) verification can be extremely time-consuming and computationally expensive due to the massive non-linearity in activation functions like ReLU. For ReLU networks under input perturbations, the ReLU neurons whose inputs may cross zero (referred to as ``unstable'' neurons) are often the key factor to determine the difficulty of verification. 
Current solutions like linear relaxations~\citep{zhang2018efficient,singh2019abstract}, semidefinite relaxations~\citep{raghunathan2018semidefinite} or Branch and Bound (BaB)~\citep{bunel2017unified,wang2021beta} suffer from either trivially loose bounds or exponentially increased complexity for large-scale networks. ($ii$) Many certified robust training methods~\citep{mirman2018differentiable,gowal2018effectiveness,wong2018scaling,balunovic2019adversarial} tend to reduce non-linearity ( ``unstable'' ReLU neurons) to improve certifiable robustness.
For example, many certified defense approaches tend increase the number of ``inactive'' neurons (ReLU neurons with negative inputs and zero output) for tightening bounds~\citep{shi2021fast}. 
This often leads to reduced standard accuracy, and in order to stably train a network with certified defense, a much longer training schedule is often required: e.g.,  \citet{xu2020automatic} used $2,000$ epochs and \citet{zhang2019towardsstable} used $3,200$ epochs for training a convolutional neural network on CIFAR-10 to state-of-the-art certified performance, much longer than conventionally adversarial training ($100\sim200$ epochs)~\citep{madry2018towards}.

\vspace{-1mm}
To address these pain points, this paper proposes a brand new alternative, \textit{linearity grafting}, to remarkably alleviate the burden of certification, by strategically operating neurons to control an appropriate level of non-linearity on a pretrained network.
Linearity grafting is inspired by neuron pruning, which strategically removes a large portion of neurons in a network while maintaining its performance. However since we aim to enhance certifiability rather than reducing network size or computation, we can \emph{relax the typical pruning paradigm} by exploiting the fact that certification is easy for linear neurons.
Specifically, linearity grafting $\underline{\mathrm{first}}$ treats a pre-trained model using adversarial training as the starting point which has good empirical robustness but usually undergoes poor certifiable robustness. $\underline{\mathrm{Then}}$, it replaces unstable yet insignificant ReLU neurons with a linear activation function, balancing the certification scalability and network expressiveness. $\underline{\mathrm{Lastly}}$, the slope and intercept are further tuned to maintain competitive classification performance thanks to its enhanced representation power compared to inactive (completely pruned) ReLU neurons. Note that vanilla neuron pruning is a \underline{special case} of our proposal, where it grafts a fixed linear function of zero to unstable neurons, yet often ends up with worse performance due to over-restriction.
Our contributions can be summarized as follows:
\vspace{-3mm}
\begin{itemize}
    \item We pioneer a thorough investigation to reveal that introducing appropriate linearity benefits certifiable robustness in multiple aspects: ($1$) substantially shrinking the bounds and reducing the number of unstable ReLU neurons; ($2$) improving certification scalability and enabling complete verification on large-scale networks; ($3$) obtaining competitive certified accuracy in a more efficient manner without explicitly certified training. 
    \vspace{-1mm}
    \item We propose a new algorithm, \textit{grafting}, for trimming down the unnecessary non-linearity for networks under verification. It linearizes the unstable yet insignificant neurons by replacing the ReLU with a linear activation function, whose slope and intercept are subsequently optimized to achieve better model performance. \vspace{-1mm}
    \item Extensive experiments across diverse network architectures on MNIST, SVHN, and CIFAR-10, validate our proposal by demonstrating the superior certified accuracies (up to $82.30\%$ improvements) \textit{without certified robust training}. Furthermore, our grafted large-scale networks with $\ge 17 \mathrm{M}$ parameters reach competitive certified accuracy ($32.30\%$ at $\epsilon=8/255$) without using certified defense training.
\end{itemize}

\section{Related Work}

\paragraph{Network Verification.} 
Neural network verification aims to formally prove or disprove desired properties of a neural network, and it becomes increasingly important to safety-crucial scenarios~\citep{katz2017reluplex,katz2019marabou}. Given sufficient time, a \emph{complete} verifier gives a definite ``yes/no" answer; on the other hand, an incomplete verifier solves a relaxed problem and may produce ``don't know''. Complete verification generally produces tighter bounds at the expense of more computations, compared to its incomplete counterpart.

Early complete verifiers are limited to very small-scale problems and relied on costly solvers like MILP~\cite{tjeng2017evaluating,dutta2017output} or SMT~\citep{katz2017reluplex,ehlers2017formal,huang2017safety}. On the other hand, branch and bound (BaB) based complete verifiers use an incomplete verifier as a sub-procedure and conduct branching on ReLU neurons~\citep{bunel2017unified,wang2018efficient,lu2019neural,botoeva2020efficient} or model input~\citep{wang2018formal,rubies2019fast,anderson2019optimization,bunel2017unified}. Recently, a series of effective BaB verifiers are proposed to use efficient iterative solvers or bound propagation methods on GPUs as the alternative for LP solvers. For example, BaDNB~\citep{de2021improved} designs a new filtered smart branching (FSB) and combines it with Lagrangian decomposition~\citep{bunel2020lagrangian} to enhance verification performance. \citet{xu2020fast} use an optimizable bound propagation method ($\alpha$-CROWN) as a massively paralleled incomplete solver on GPUs and an LP solver for completeness checking. $\beta$-CROWN~\citep{wang2021beta} completely eliminate the dependency on an LP solver with optimizable constraints, greatly boosting verification performance and efficiency.

An incomplete verifier is usually weaker but faster than complete verifiers. They typically rely on duality~\citep{dvijotham2018dual,wong2018provable}, linear relaxations~\citep{wang2018efficient,zhang2018efficient,salman2019convex} or semidefinite relaxations~\citep{raghunathan2018semidefinite,dathathri2020enabling}. Cheap incomplete verifiers can be applied at training time~\citep{wong2018provable,wang2018mixtrain,mirman2018differentiable,wong2018scaling,gowal2018effectiveness,balunovic2019adversarial,shi2021fast} for certified defense mechanisms. Our work achieves certified robustness without relying on a certified defense, but with the help of a more powerful complete verifier on an adversarially trained model with post-training linearity grafting.

\vspace{-3mm}
\paragraph{Pruning.} Pruning~\citep{lecun1990optimal,han2015deep} as one of the effective compression approaches eliminates the redundancy in over-parameterized neural networks, achieving storage and computational savings with nearly unimpaired performance. According to the granularity of removed components, pruning can be roughly categorized into two groups: ($1$) Unstructured pruning, which zeroes out insignificant parameters based on some heuristics like weight magnitude~\citep{han2015deep,han2015learning}, gradient~\citep{molchanov2019importance}, or hessian~\citep{lecun1990optimal} statistics. 
It usually leads to competitive performance but is hard to be accelerated due to irregular sparsities. 
($2$) Structural pruning, which discards sub-structures (e.g., channels~\citep{liu2017learning,zhou2016less,he2017channel} or layers~\citep{wang2018skipnet, wu2018blockdrop, zhang2019all}) based on optimized importance scores or heuristics. It produces hardware-friendly subnetworks at the cost of moderate performance degradation. Normally, structural pruning also brings the reduction of intermediate activation (or neuron) dimension. \citet{dhillon2018stochastic,ye2020accelerating} particularly devote themselves to trimming down the number of neurons for empirical robustness and network acceleration. 

\vspace{-3mm}
\paragraph{Pruning and Robustness.} On one hand, \citet{gui2019model,ye2019adversarial,sehwag2019towards,jordao2021effect,fu2021drawing} pursue empirical robust and efficient subnetworks that can be deployed in security-critical yet resource-limited platforms. \citet{sehwag2020hydra} also consider the trade-off between certified robustness and efficiency by learning sparse weights. On the other hand, \citet{wang2018defending,gao2017deepcloak,dhillon2018stochastic} treat model compression as an empirical defense mechanism. \citet{xiao2018training} for the first time utilize \textit{weight} sparsity to speed up the certification of multi-layer perceptron (MLP) with LP or MILP verifiers. However, sparse weights do not necessarily result in zero or stable activation, especially for convolutional neural networks. \citet{han2021scalecert} conduct certified defense against adversarial patches by removing related superficial neurons.

Our work is fundamentally different from existing literature by recognizing \textbf{two important facts}: \textit{(i) linearity plays a central role in the difficulty of robustness certification; and (ii) sparsity is only a special case of linearity}. Our proposed grafting focuses on linearizing unstable and insignificant neurons. Thus, it directly reduces ReLU instability, obtains tight verification bound, and in the meanwhile ensures sufficient flexibility for restoring model performance. As a result, we improve robustness verification in terms of certified robust accuracy, speed, and scalability.

\vspace{-0.5mm}
\section{Preliminaries}

\vspace{-0.5mm}
\subsection{The Neural Network Verification Problem}

Let $x\in\mathbb{R}^{d_0}$ be the input sample of a $L$-layer deep neural network $f:\mathbb{R}^{d_0}\rightarrow\mathbb{R}$, and $f$ is parameterized by weights $\mathbf{W}^{(i)}\in\mathbb{R}^{d_i\times d_{i-1}}$ and biases $\mathbf{b}^{(i)}\in\mathbb{R}^{d_i}$, $i\in\{1,\cdots,L\}$, respectively. Then, we have $f(x)=z^{(L)}(x)$, $z^{(i)}(x)=\mathbf{W}^{(i)}\hat{z}^{(i-1)}(x)+\mathbf{b}^{(i)}$, $\hat{z}^{(i)}(x)=\mathcal{A}(z^{(i)}(x))$, and $\hat{z}^{(0)}(x)=x$, where $\mathcal{A}$ is activation function, $z^{(i)}_j$ and $\hat{z}^{(i)}_j$ represents the pre-activation and post-activation values of the $j$-the neuron in the $i$-th layer. The neural network verification can be formulated as:
\vspace{-1mm}
{\begin{align}\label{eq:verification}
\begin{array}{ll}
\displaystyle & \mathrm{min}f(x):=z^{(L)}(x)\\
& \mathbf{s.t.}\ z^{(i)}=\mathbf{W}^{(i)}\hat{z}^{(i-1)}+\mathbf{b}^{(i)},\hat{z}^{(i)}=\mathcal{A}(z^{(i)})\\
& \hat{z}^{(0)}=x, \ x\in \mathcal{C}
\end{array}
\end{align}}%
where the set $\mathcal{C}$ denotes the allowed input region, and the goal is to find the lower bound of $f(x)$ given $x\in\mathcal{C}$. Without loss of generality, $\mathcal{C}$ can be an $\ell_{\infty}$ ball with a radius of $\epsilon$ around a data example $x_0$, i.e., $\mathcal{C}=\{x|\|x-x_0\|_{\infty}\le\epsilon\}$. We consider the canonical specification $f(x)>0$: if we can prove that $f(x)>0$, $\forall x\in\mathcal{C}$, we say $f(x)$ is verified. Other ``specifications'' like the margin between logits in practical settings can be often seen as an output layer of networks, merged into $f(x)$ during verification, and turned back to the canonical specification~\citep{wang2021beta}.

\vspace{-1mm}
\paragraph{Unstable neurons.} The most challenging part of verification is the non-linearity of activation functions $\mathcal{A}$. Specifically, given $\mathcal{A}(z_j^{(i)}):=\mathrm{ReLU}(z_j^{(i)})=\mathrm{max}(0,z_j^{(i)})$, we denote its intermediate layer bounds as $\mathbf{l}_j^{(i)}\le z_j^{(i)}\le\mathbf{u}_j^{(i)}$, which bound the input of this specific ReLU neurons given $x \in \mathcal{C}$. Given intermediate layer bounds, each ReLU neuron (i.e., activation value) can be grouped into two categories: ($1$) if $\mathbf{l}_j^{(i)}\ge0$ or $\mathbf{u}_j^{(i)}\le0$, ReLU lies in linear active ($\hat{z}_j^{(i)}=z_j^{(i)}$) or inactive $(\hat{z}_j^{(i)}=0)$ regions ; ($2$) if $\mathbf{l}_j^{(i)}\le 0\le\mathbf{u}_j^{(i)}$, we call this ReLU neuron as an \textit{unstable neuron}, which usually places obstacles to certification.

\subsection{Branch and Bound based Complete Verifier}

A large portion of complete verifiers~\citep{bunel2017unified,henriksen2020efficient,wang2021beta} adopt the Branch and Bound (BaB) method. They first divide the domain of the verification problem $\mathcal{C}$ into two subdomains $\mathcal{C}_1=\{x\in\mathcal{C},z_j^{(i)}\ge0\}$ and $\mathcal{C}_2=\{x\in\mathcal{C},z_j^{(i)}\leq0\}$, according to an unstable ReLU neuron $z^{(i)}_j$ that now becomes linear in each subdomain. Then, cheap incomplete verifiers are applied to estimate the \emph{lower bound} of the objective~\eqref{eq:verification}. for each subdomain under some relaxations. If subdomain $\mathcal{C}_i$'s lower bound is greater than $0$, $\mathcal{C}_i$ is verified; otherwise, another unstable ReLU neuron will be further split over domain $\mathcal{C}_i$, until all subdomains are verified. Therefore, the amount of non-linearity or unstable neurons directly determines the computation and memory complexity of complete verifiers. When each subdomain is solved using an incomplete verifier, the tightness of the lower bound also heavily depends on the number of unstable ReLU neurons - too many unstable neurons may lead to vacuous bounds and zero verified accuracy. Thus, the goal of our grafting linearity is to scale up verification by reducing unstable ReLU neurons that are insignificant for model clean performance.

\begin{figure*}[t]
\centering
\includegraphics[width=1\linewidth]{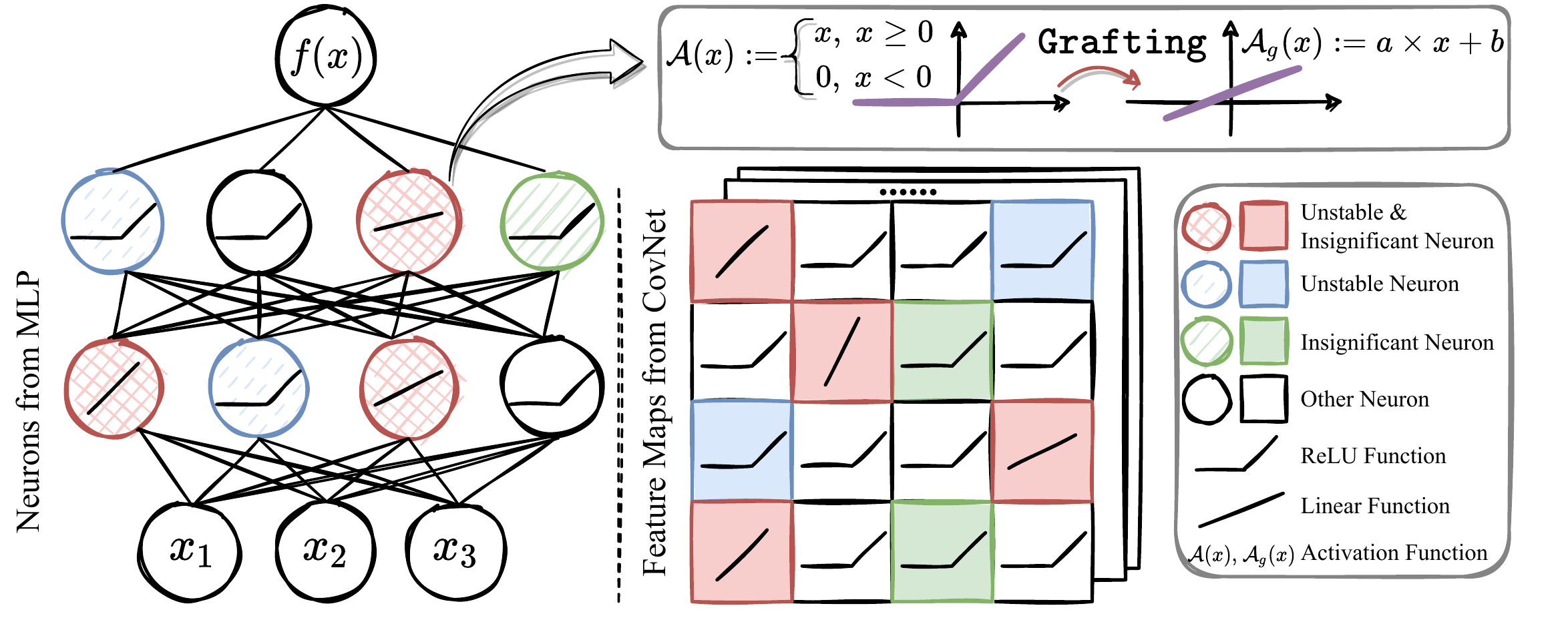}
\vspace{-6mm}
\caption{The overall illustration of our proposed \texttt{\textbf{linearity grafting}}, which linearizes the piece-wise activation function of unstable and insignificant neurons with a linear function (i.e., $a\times x+b$). (\textit{Left}) and (\textit{Right}) show demos of applying grafting to MLP and ConvNet, where unstable \& insignificant neurons (e.g., \textcolor{Maroon}{red} cycles or blocks) are our main focus.}
\vspace{-2mm}
\label{fig:methods}
\end{figure*}

\section{Grafting for Certified Verification}
\label{sec:grafting}

This section presents the detailed processes of linearity grafting: ($i$) starting from an empirical robust DNN; ($ii$) identifying and linearizing the unstable \& insignificant neurons; ($iii$) further optimizing the parameters in grafted linear activation functions; ($iv$) performing complete verification on grafted NNs. We detail each step below.

\begin{figure}[t]
\centering
\includegraphics[width=1\linewidth]{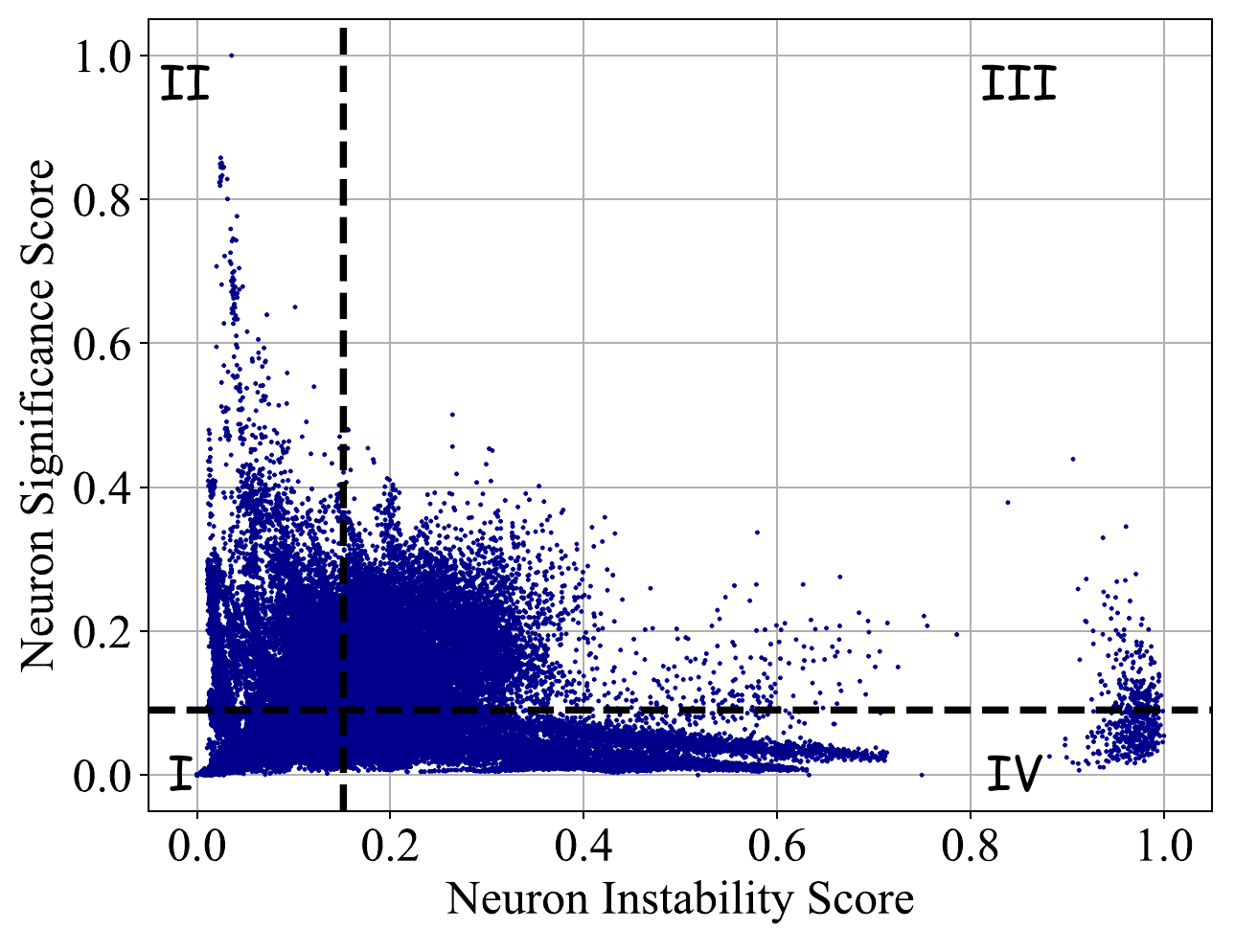}
\vspace{-8mm}
\caption{Normalized neurons' significance scores ($r^{(i)}_s$) over instability scores ($r^{(i)}_u$). Significance scores are calculated based on the absolute magnitude of activation gradients, and instability scores are computed by a SOTA verifier, $\alpha\!,\beta$-CROWN. Intuitively, regions (I, II, III, IV) indicate (\textit{insignificant and stable, significant and stable, significant and unstable, insignificant and unstable}) neurons.}
\vspace{-2mm}
\label{fig:scores}
\end{figure}

\textbf{$\rhd$ Robustify a DNN using adversarial training as the starting point.} We start from an empirical robust model from cheap adversarial defenses, which might not be certifiable especially when the model is large. We aim to improve its certifiable robustness without explicit certified defense robust training, which is usually slow and can hurt model performance. Specifically, a fast adversarial training approach~\citep{andriushchenko2020understanding} with gradient alignment regularization is adopted in our case, whose training loss is:
\begin{align}\label{eq:our_mix}
\begin{array}{ll}
\displaystyle \mathcal{L}:=\mathcal{L}_{\mathrm{FAT}} + \lambda\times\mathcal{R}_{\mathrm{GA}},
\end{array}
\end{align}
where $\mathcal{L}_{\mathrm{FAT}}$ is the loss of fast adversarial training~\citep{wong2020fast} with GradAlign regularizer $\mathcal{R}_{\mathrm{GA}}$; $\lambda$ controls the portion of regularization, which is $0.2$ in our case following the default choice in~\citet{andriushchenko2020understanding}.  


\textbf{$\rhd$ Identify insignificant and unstable neurons.} 
Given a trained DNN, our goal is to reduce the ReLU instability (i.e., unstable neurons), gain certifiable robustness, and preserve sound generalization. To locate qualified neuron candidates, we design the selection pipeline as below:

\ding{182} Rank all neurons according to the number of images for which it is unstable, normalize the rank and treat it as the neuron instability 
score $r^{(i)}_u\in[0,1]$, where $i$ is the neuron index. 

\vspace{-1mm}
\ding{183} Compute the importance of each neuron via certain heuristics or optimized scores (e.g., the magnitude of activation's gradient), rank and normalize them to obtain the neuron significance score $r^{(i)}_s\in[0,1]$, where $i$ is the neuron index.

\vspace{-1mm}
\ding{184} Based on some criteria like $\mathrm{argmax}_i \gamma\times r^{(i)}_u-r^{(i)}_s$ to identify the insignificant and unstable neurons, where $\gamma$ adjusts the portion of instability scores. If $\gamma\to 0$ or $\infty$, it chooses neurons purely on significant or instability scores. In our case, we decay $\gamma$ from $2$ to $0$, along with the total number of grafted neurons. For example, to graft $20\%$ neurons, we select every $5\%$ neurons with $\gamma=2\to1.5\to1\to0$, respectively.

\begin{figure*}[t]
\centering
\includegraphics[width=1\linewidth]{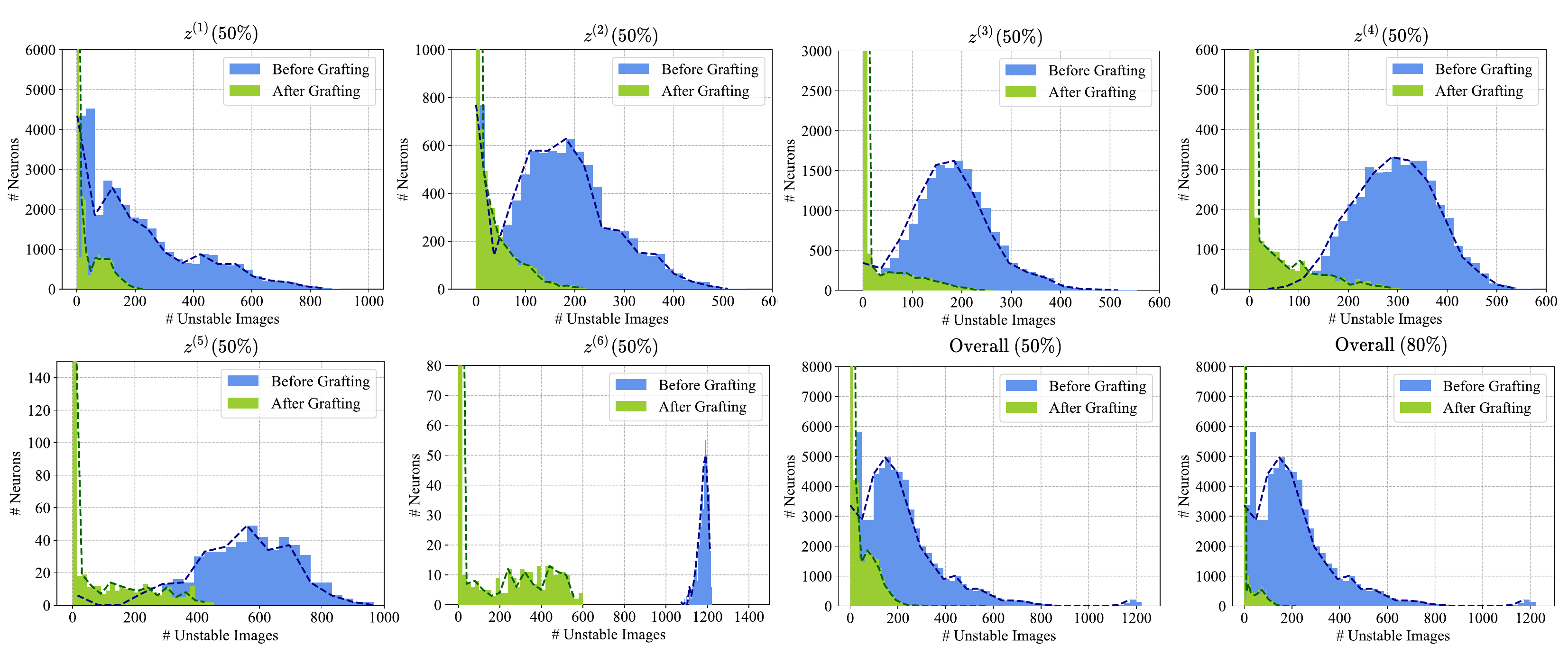}
\vspace{-9mm}
\caption{Layer-wise ($z^{(i)}$) and overall unstable neuron distribution of the $7$-layer ConvBig on CIFAR-10, before and after performing grafting on $50\%$ or $80\%$ neurons. On these figures, a bar located at $m$ unstable images (x-axis) with a height of $n$ neurons (y-axis) means that $n$ neurons are unstable for $m$ images in the test set.}
\vspace{-3mm}
\label{fig:unstable_neurons}
\end{figure*}

\begin{figure}[t]
\centering
\includegraphics[width=1\linewidth]{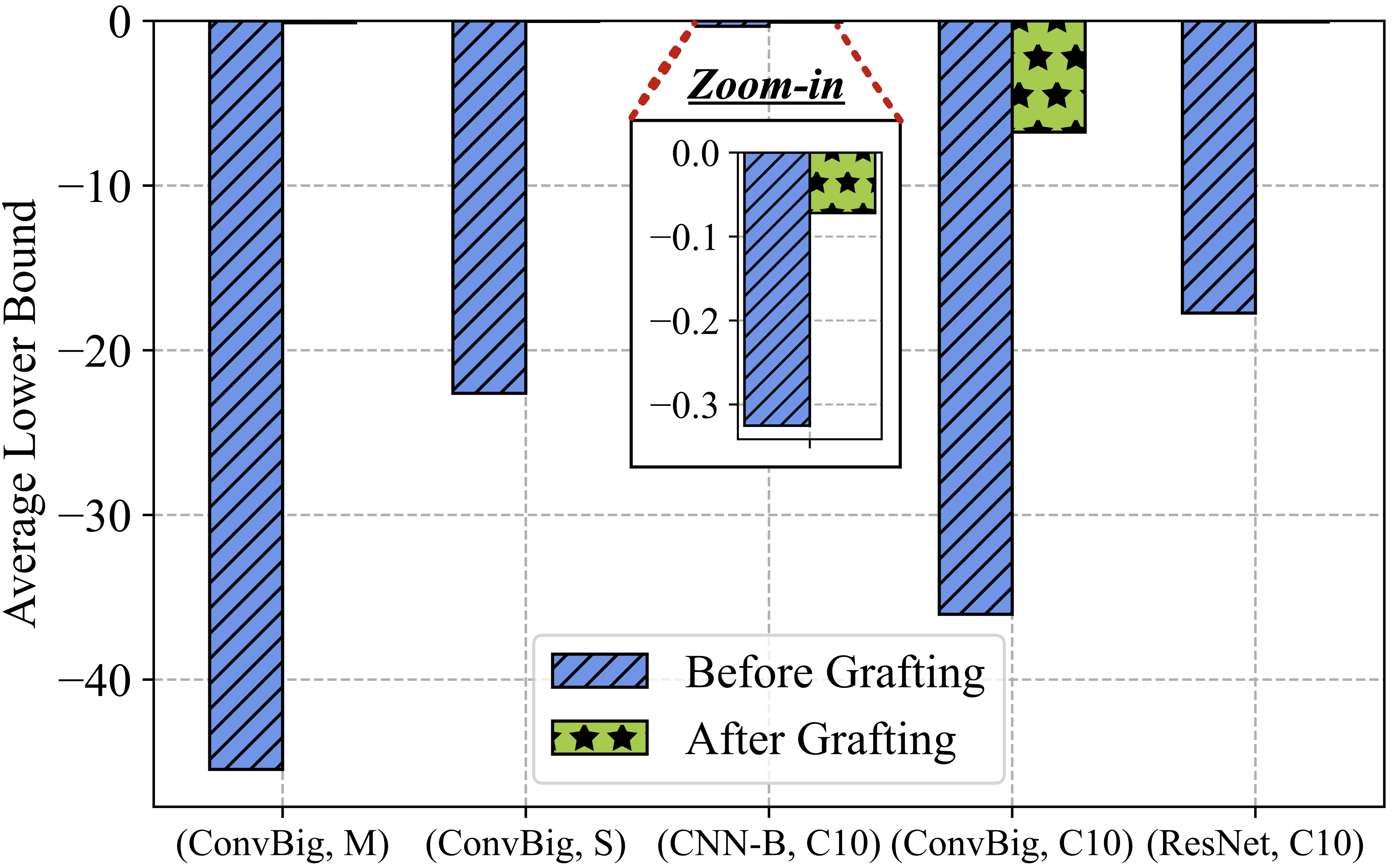}
\vspace{-8mm}
\caption{Average lower bounds after BaB branching of verification~\citep{wang2021beta} for models before and after grafting $50\%$ neurons on 5 models. Bounds are produced by $\alpha\!,\beta$-CROWN. A very negative lower bound often means the bound is vacuous due to the large amounts of unstable neurons.}
\vspace{-5mm}
\label{fig:bound}
\end{figure}

Figure~\ref{fig:scores} demonstrates the distributions of $r^{(i)}_u$ and $r^{(i)}_s$, which can be roughly divided into four regions (I, II, III, IV), and each region has a quarter of neurons. Region IV is our main focus, implying insignificant and unstable neurons. An investigation of different heuristics for computing importance scores is provided in Sec.~\ref{sec:extra_study}.

\textbf{$\rhd$ Linearize and tune the grafted activation functions.} As the last step of linearity grafting, we linearize the selected neurons (i.e., \textcolor{Maroon}{red} cycles or blocks in Figure~\ref{fig:methods}) by replacing the non-linear activation function (e.g., ReLU $\mathcal{A}(x)$) with a parameterized linear function $\mathcal{A}_g(x):=a\times x+b$, where $a$ and $b$ are trainable slope and intercept respectively. Note that each neuron has its own associated $a$ and $b$. Since the exorbitant non-linearity (or unstable ReLU neurons) causes burdensome computation and limits the scalability of verification, our grafting directly discards redundant non-linearity and injects learnable linearity, encouraging DNN to be more amenable for verification. Furthermore, contrary to typical certified defense where the inactive neurons may dominate in the network~\citep{shi2021fast}, the grafted neurons are never forced to be inactive and can be optimized towards better model performance.   

\textbf{$\rhd$ Robustness verification with a complete verifier.} After linearity grafting, the robustness of our model is verified using a complete verifier. Unlike cheap verification methods used in certified defense such as IBP~\citep{gowal2018effectiveness,mirman2018differentiable} where the models have to be trained to adapt to a very weak certification method which may impose strong constraints on model expressiveness, 
a complete verifier is much more powerful and allows the model to be more flexible for a better trade-off between verified accuracy and standard accuracy. We choose the a state-of-the-art (SOTA) complete verifier, $\alpha\!,\beta$-CROWN~\citep{zhang2018efficient,xu2020automatic,wang2021beta}, as our default certification method, with added support on a customized activation function that can be selected from either a ReLU function (neurons not grafted) or a linear function (grafted neurons). We find that SOTA complete verifiers can produce quite competitive results when an sufficient number of unstable neurons are grafted, as we will show in Section~\ref{sec:exp}.



\definecolor{LightCyan}{rgb}{0.88,1,1}
\newcolumntype{a}{>{\columncolor{LightCyan}}c}

\begin{table*}[t]
\centering
\caption{\textbf{Unstable neuron ratio (UNR $\%$), verified accuracy (\colorbox{LightCyan}{VA} $\%$), standard accuracy (SA $\%$), PGD-$100$ robust accuracy (RA $\%$), and average time (s)} of FAT trained models w./w.o. grafting on MNIST, SVHN, and CIFAR-10. $\alpha\!,\!\beta$-CROWN, a SOTA complete verifier is used to compute VA. The target $\ell_\infty$ norm perturbation is $\epsilon=\frac{2}{255}$ except for MNIST. ``OOM" indicates that DNNs have too many unstable neurons and the verifier is unable to load it with 48 GB GPU memory, leading to ``$\infty$" verification time and a null VA (``-").}
\resizebox{1\linewidth}{!}{
\begin{tabular}{@{}l|caccc|caccc|caccc}
\toprule
\multicolumn{1}{c|}{\multirow{2}[2]{*}{FAT ($\epsilon=\frac{2}{255}$)}} & \multicolumn{5}{c|}{(ConvBig, MNIST w. $\epsilon=0.1$)} & \multicolumn{5}{c|}{(ConvBig, SVHN)} & \multicolumn{5}{c}{(CNN-B, CIFAR-10)}  \\ \cmidrule{2-16}
\multicolumn{1}{c|}{} & UNR & VA & SA & RA & Time & UNR & VA & SA & RA & Time & UNR & VA & SA & RA & Time  \\ \midrule
Baseline & 31.27 & 0.10 & 99.29 & 97.14 & 262.11 & 10.78 & 16.70 & 89.71 & 75.74 & 218.49 & 15.85 & 37.40 & 79.95 & 62.23 & 127.50 \\ \midrule
SAP~\citep{dhillon2018stochastic} ($50\%$)& 7.38 & 4.20 & 99.22 & 96.34 & 292.94 & 5.65 & 25.90 & 89.85 & 76.03 & 195.87 & 6.27 & 47.30 & 75.10 & 58.01 & 58.98 \\
GAP$^\dagger$~\citep{ye2020accelerating} ($50\%$) & 17.29 & 3.50 & 99.19 & 96.46 & 295.21 & 6.14 & 26.20 & 90.09 & 77.28 & 195.78 & 10.22 & 42.50 & 79.05 & 61.81 & 103.03 \\
Hydra$^\ddagger$~\citep{sehwag2020hydra} ($50\%$) &15.39 &12.70 & 98.90 & 95.22 & 269.71 & 5.04 & 26.60 & 81.28 & 62.92 & 172.98 & 6.28 & 44.40 & 72.99 & 55.55 & 59.99 \\ \midrule
Random Grafting ($50\%$) &17.16 &12.00 & 98.93 & 95.38 &273.94 & 6.13 & 37.40 & 87.37 & 73.27 & 150.23 & 9.07 & 42.50 & 75.02 & 57.19 & 83.25  \\
Grafting ($50\%$) & 5.85 & 82.30 & 98.68 & 92.73 & 40.21 & 3.11 & 57.80 & 78.75 & 63.90 & 16.68 & 5.36 & 50.40 & 74.08 & 58.76 & 39.32 \\ \midrule
Grafting ($30\%$) & 10.43 & 59.40 & 99.13 & 95.24 & 137.40 & 5.45 & 56.80 & 80.71 & 66.05 & 31.76 & 7.15 & 49.00 & 77.10 & 60.87 & 64.80 \\
Grafting ($80\%$) & 4.04 & 82.40 & 98.63 & 92.71 & 39.64 & 1.63 & 58.70 & 78.56 & 63.91 & 12.93 & 1.87 & 44.40 & 61.20 & 48.34 & 15.25 \\
\toprule
\multicolumn{1}{c|}{\multirow{2}[2]{*}{FAT ($\epsilon=\frac{2}{255}$)}} & \multicolumn{5}{c|}{(ResNet-4B, CIFAR-10)} & \multicolumn{5}{c|}{(ConvBig, CIFAR-10)} & \multicolumn{5}{c}{(ConvHuge, CIFAR-10)} \\ \cmidrule{2-16}
\multicolumn{1}{c|}{} & UNR & VA & SA & RA & Time & UNR & VA & SA & RA & Time & UNR & VA & SA & RA & Time \\ \midrule
Baseline & 19.94 & 0.80 & 76.69 & 60.14 & 45.56 & 17.75 & 1.30 & 84.90 & 68.10 & 121.61 & OOM & - & 90.68 & 73.57  & $\infty$ \\ \midrule
SAP~\citep{dhillon2018stochastic} ($50\%$) & 6.18 & 21.70 & 49.03 & 38.30 & 137.77 & 5.54 & 25.80 & 65.08 & 50.45 & 156.28 & 8.52 &2.00 & 80.29 & 60.29 & 181.06 \\
GAP$^\dagger$~\citep{ye2020accelerating} ($50\%$) & 13.67 & 5.10 & 68.42 & 53.43 & 239.14 & 10.97 & 1.10 & 81.91 & 64.50 & 190.42 &7.43 &1.00 & 86.38 & 67.91 &111.77 \\
Hydra$^\ddagger$~\citep{sehwag2020hydra} ($50\%$) & 9.52 & 15.10 & 42.01 & 31.27 & 162.34 & 11.10 & 1.10 & 67.97 & 47.77 & 297.19 &9.88 &1.00 & 70.68 & 48.81 &291.00 \\ \midrule
Random Grafting ($50\%$)& 13.59  & 7.40 & 69.56 & 52.53 & 267.74 & 12.23 & 3.90 & 79.33 & 60.92 & 285.71 &11.34 &1.00  & 84.47 & 64.76 &206.97 \\
Grafting ($50\%$) & 6.03 & 38.10 & 60.13 & 46.12 & 42.83 & 4.32 & 39.12 & 62.23 & 47.73 & 42.80 & 4.41 & 28.30 & 62.62 & 49.37 & 155.78  \\ \midrule
Grafting ($30\%$) & 12.89 & 24.50 & 63.71 & 49.16 & 153.69 & 10.30 & 27.30 & 71.97 & 54.97 & 159.74 & OOM & - & 90.19 & 72.34 & $\infty$ \\
Grafting ($80\%$) & 2.91 & 39.70 & 57.64 & 44.61 & 25.16 & 1.89 & 41.00 & 55.20 & 44.27 & 10.87 & 0.17 & 32.30 & 40.80 & 33.43 & 4.06   \\
\bottomrule
\end{tabular}}
\label{tab:main_res_2}%
\begin{tablenotes}
\scriptsize
\item $\dagger$ The heuristic of activation gradient magnitude~\citep{ye2020accelerating} is utilized to guide activation pruning. 
\item $\ddagger$ Based on the official implementation of~\citet{sehwag2020hydra}, we extend the original sparse mask learning to activation sparsification. 
\end{tablenotes}
\vspace{-3mm}
\end{table*}%

\section{Experiments}
\label{sec:exp}

\paragraph{Datasets and architectures.} Our experiments are conducted on three representative datasets in adversarial robustness and verification literature, MNIST~\citep{deng2012mnist}, SVHN~\citep{netzer2011reading} and CIFAR-10~\citep{krizhevsky2009learning}. For MNSIT and SVHN, we consider a 7-layer convolutional neural network (ConvBig) from~\citet{mirman2018differentiable,singh2019abstract}. For CIFAR-10, we adopt four network architectures, including a 4-layer CNN (CNN-B)~\citep{dathathri2020enabling}, a 11-layer ResNet-4B~\citep{bak2021second}, ConvBig~\citep{mirman2018differentiable}, and a wider 7-layer CNN (ConvHuge) with $17.2$M parameters. Note that it is usually impractical to verify a purely adversarially trained model with over $10$M parameters, and our linearity grafting procedure enables complete verification for these models.

\vspace{-1mm}
\paragraph{Training details.} For fast adversarial training~\cite{wong2020fast}, we adopt the effective GradAlign regularization~\citep{andriushchenko2020understanding} with a coefficient of $0.2$, for all $200$ training epochs. The learning rate starts from $0.1$ and decays by ten times at epochs $100$ and $150$, while the batch size is $128$. We use an SGD optimizer with $0.9$ momentum and $5\times10^{-4}$ weight decay. 
During the fine-tuning of grafted networks,
an initial learning rate of $0.01$ is used for trainable slopes and intercept $(a,b)$ of grafted neurons, and $0.001$ for original model parameters. And the learning rate decays with a cosine annealing schedule of $100$ training epochs. Other configurations are the same as the corresponding original pre-trained networks.

\vspace{-1mm}
\paragraph{Evaluation metrics.} We evaluate our methods on the official test sets with five classical metrics: ($1$) unstable neuron ratio (\textbf{UNR} $\%$) is number of unstable neurons divided by total number of neurons; ($2$) verified accuracy (\textbf{\colorbox{LightCyan}{VA}} $\%$) is the percentage of verifiably robust test images; ($3$) standard accuracy (\textbf{SA} $\%$); ($4$) robust accuracy (\textbf{RA} $\%$) is the percentage of robust test images under empirical attacks; and ($5$) average verification time per sample (\textbf{Time}). RA is calculated on perturbed testing samples generated by PGD-$100$~\citep{madry2019deep} with $100$ restarts. VA, Time, and UNR are computed via the current SOTA complete verifier $\alpha\!,\!\beta$-CROWN with a time-out of 300 seconds and other parameters leaving at the default. If the time of complete verification for an image exceeds $300$s, its certification fails. SA, RA are evaluated on the full test set. VA is computed on the first $1,000$ images due to its high computational cost, following the setup in~\citet{NEURIPS2019_0a9fdbb1,muller2021prima,tjandraatmadja2020convex,wang2021beta}. The reported verification time excludes examples that are classified incorrectly or attacked successfully. 

\vspace{-1mm}
\subsection{The Superiority of Grafting for Verification} 

In this section, we compare \textit{grafting} with five pruning baseline methods: ($i$) \textit{Baseline} without neuron pruning or grafting; ($ii$) \textit{Stochastic Activation Pruning (SAP)}~\citep{dhillon2018stochastic} that removes the activation with the least magnitude; ($iii$) \textit{Gradient-based Activation Pruning (GAP)} which leverages activation's gradient~\citep{ye2020accelerating} as a heuristic to prune neurons; ($iv$) \textit{Hydra}~\citep{sehwag2020hydra} that learns sparse activation masks; ($v$) \textit{Random Grafting} as a sanity check which randomly selects neurons for grafting linearity, without considering the ranking discussed in Section~\ref{sec:grafting}.

\vspace{-4mm}
\paragraph{Improved verification performance.} Experimental results with target perturbation radius $\epsilon=\frac{2}{255}$ and $\epsilon=\frac{8}{255}$ are collected in Table~\ref{tab:main_res_2}, and Table~\ref{tab:main_res_8} (in Appendix), respectively. Several consistent observations can be drown from these extensive evaluations with four network architectures on MNIST (M), SVHN (S) and CIFAR-10 (C10) datasets: 

\ding{182} Compared to the baseline, applying grafting to $50\%$ neurons of \{(ConvBig, M), (ConvBig, S), (CNN-B, C10), (ResNet, C10), (ConvBig, C10)\} achieves \{$82.20\%$, $41.10\%$, $13.00\%$, $37.30\%$, $37.82\%$\} certified accuracy boosts and \{$25.42\%$, $7.67\%$, $10.49\%$, $13.91\%$, $13.43\%$\} unstable neuron ratio reductions, where models have never been explicitly trained using a certified defense. Such impressive VA improvements evidence the effectiveness of grafted linearity, which balances the DNN expressiveness and certifiable robustness by controlling the number of unstable neurons. Although it comes with \{$0.61\%$, $10.96\%$, $5.87\%$, $16.56\%$, $22.67\%$\} SA drops, grafted NNs still reach a clearly better trade-off, especially for (ConvBig, M / S). Conclusions are also valid in the more challenging case of $\epsilon=\frac{8}{255}$ in Table~\ref{tab:main_res_8}.

\ding{183} Compared to existing activation pruning mechanisms, grafting demonstrates a substantial advantage in terms of reduced UNR, enhanced VA, and overall SA \& VA trade-offs. It echos our intuition since grafting learnable linear activation functions is a more general and powerful solution than directly replacing ReLUs by zero functions (i.e., neuron pruning). Among three pruning baselines on CIFAR-10, SAP has the best VA and inferior SA; GAP reaches a superior SA and the worst VA; Hydra achieves an in-between performance. 

\ding{184} We enable complete verification for a large-scale ConvHuge model with $17.2$M parameters on a single GPU, by grafting appropriate linearity. We notice that only grafting more than $50\%$ neurons can reach satisfied VA like $>28.30\%$, while it suffers moderate SA drops due to the challenging setting. The major reason for improved certification scalability is the reduced non-linearity that essentially alleviates the computation and memory bottleneck of complete verification.

\ding{185} Comparing grafted NNs with different ratios, from $30\%\to50\%$, certified accuracies are further increased with significantly diminished unstable neurons; from $50\%\to80\%$, excessive injected linearity starts to harm both generalization and robustness, particularly for CIFAR-10.

\begin{figure}[t]
\centering
\includegraphics[width=1\linewidth]{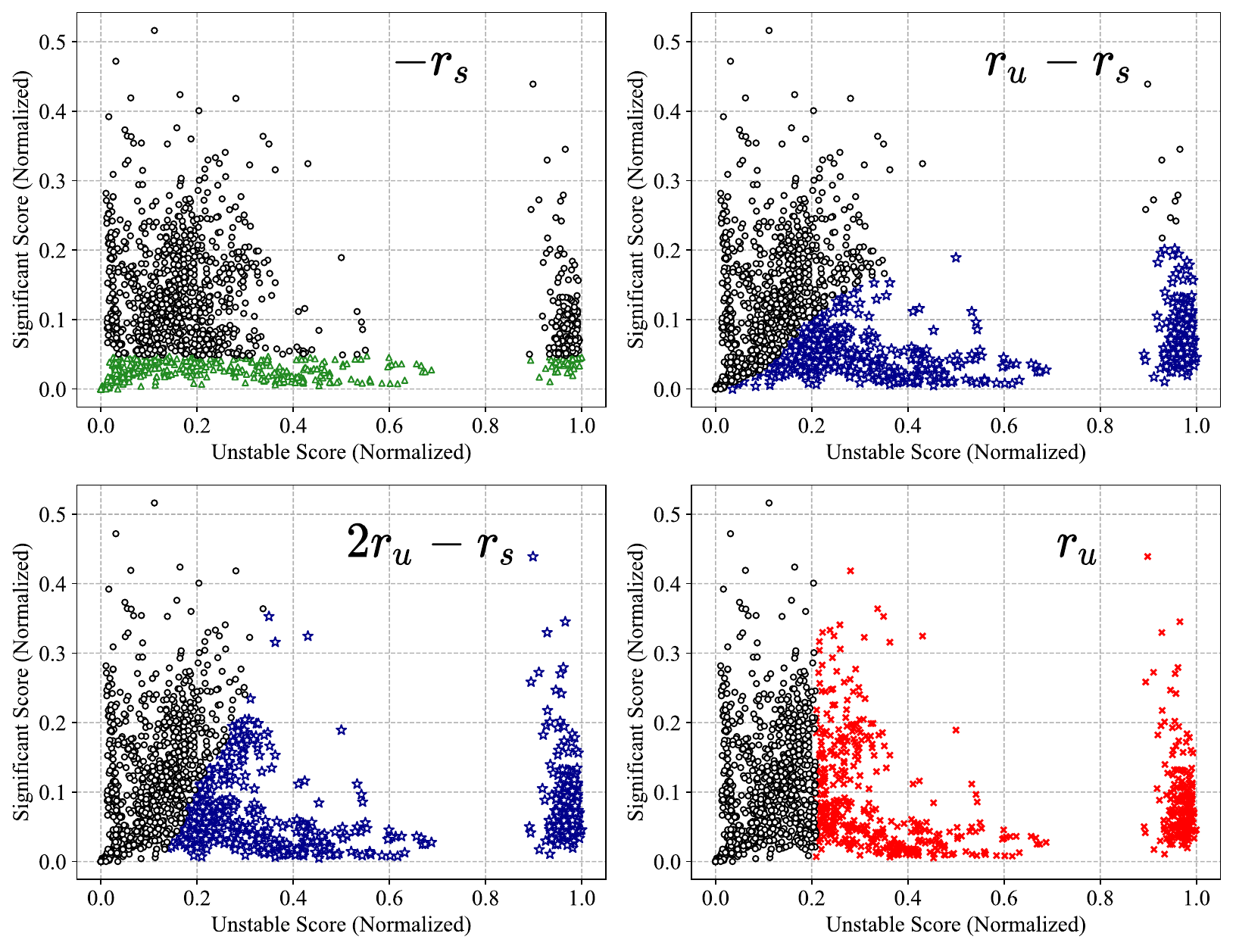}
\vspace{-9mm}
\caption{Neuron selections based on diverse picking criteria. \textcolor{ForestGreen}{$\boldsymbol{\triangle}$}, \textcolor{Blue}{$\bigstar$}, and \textcolor{Maroon}{$\boldsymbol{\Diamond}$} indicate insignificant-only, insignificant-and-unstable, and unstable-only neuron selections respectively. $r_u$ and $r_s$ are defined in Section~\ref{sec:grafting}.}
\vspace{-5mm}
\label{fig:criterion}
\end{figure}

\ding{186} Thanks to injected linearity and reduced branching in BaB, most of our grafted NNs largely cut down the complete verification time (by $5.99\%\sim 94.08\%$). Two exceptions are the $30\%$ grafted NN of (ResNet-4B / ConvBig, C10), which take a longer time than baseline networks although they have less unstable neurons. It is because the baseline model ($0.80$ / $1.30$ VA) has vacuous verification bounds for some data points (Figure~\ref{fig:bound}), which triggers an early stop in the verifier to give up early.

\vspace{-2mm}
\paragraph{Less unstable neurons and tighter bounds.} Unstable neuron distributions and verification bounds are another two necessary angles to examine certifiable robustness.

Specifically, Figure~\ref{fig:unstable_neurons} presents the layer-wise and overall unstable neuron distribution of networks before and after grafting. We find that grafting substantially pushes the distribution towards zero (i.e., the stable status), implying reduced unstable neurons after grafting. It is coherent with the findings in Figure~\ref{fig:bound} that grafted NNs consistently enjoy much tighter certified bounds (e.g., shrunk by a factor of $5\sim 744$) and are more amenable to verification, compared to their vanilla counterparts. These impressive benefits should be credited to properly inserted linearity.

\subsection{Dissecting Grafting}
\vspace{-1mm}

In this section, we perform a comprehensive analysis of each component in linearity grafting and located grafting masks for visualization.

\begin{table}[t]
\centering
\vspace{-2mm}
\caption{Ablation on grafting criterion. Unstable neuron ratio (UNR $\%$), \colorbox{LightCyan}{VA} ($\%$), SA ($\%$), and RA ($\%$) of ConvBig with $50\%$ grafted neurons on CIFAR-10 are reported.}
\resizebox{1\linewidth}{!}{
\begin{tabular}{@{}l|cacc}
\toprule
\multirow{1}[1]{*}{Grafting Criterion} & UNR & VA & SA & RA \\ \midrule
$-r_s$ & 10.35 & 2.10 & 82.35 & 64.28 \\
$r_u-r_s$ & 6.39 & 14.50 & 77.88 & 59.91 \\
$2r_u-r_s$ & 4.32 & 38.90 & 62.15 & 47.70 \\
$r_u$ & 4.13 & 38.70 & 59.39 & 45.77 \\ \midrule
$\gamma\times r_u-r_s$ ($\gamma$ linearly decays $2\to 0$) & 4.32 & 39.12 & 62.23 & 47.73 \\
\bottomrule
\end{tabular}}
\label{tab:criterion}%
\vspace{-3mm}
\end{table}%

\vspace{-2mm}
\paragraph{Grafting criterion.} Intuitively, the neuron picking criterion plays an essential role in the achievable performance of grafting. As a show case, we study four criteria variants including insignificant-only ($-r_s$)\footnote{We omit the neuron index $i$ for simplicity.}, insignificant-and-unstable ($r_u-r_s$ and $2r_u-r_s$), and unstable-only ($r_u$), and display the corresponding selected candidate neurons in Figure~\ref{fig:criterion}. We see that the criteria with a larger weighted $r_u$ tends to pick more unstable neurons. Quantitative evaluations are collected in Table~\ref{tab:criterion}. The results show that: \ding{182} grafting neurons solely based on neuron significance scores ($-r_s$) or neuron instability scores ($r_u$) leads to an inferior trade-off between SA and VA. For example, $82.35\%$ SA and $2.10\%$ VA for the criteria $-r_s$; $59.39\%$ SA and $38.70\%$ VA for the criteria $r_u$. \ding{183} The criterion involving both $r_u$ and $r_s$ reach an improved balance between network expressiveness and certifiable robustness. \ding{184} The linearly decayed $\gamma$ establishes the superior complete verification accuracy. It tends to bias on neuron significance scores when grafting more neurons, which perseveres a better network expressiveness.

\begin{figure}[!ht]
\centering
\vspace{-2mm}
\includegraphics[width=1\linewidth]{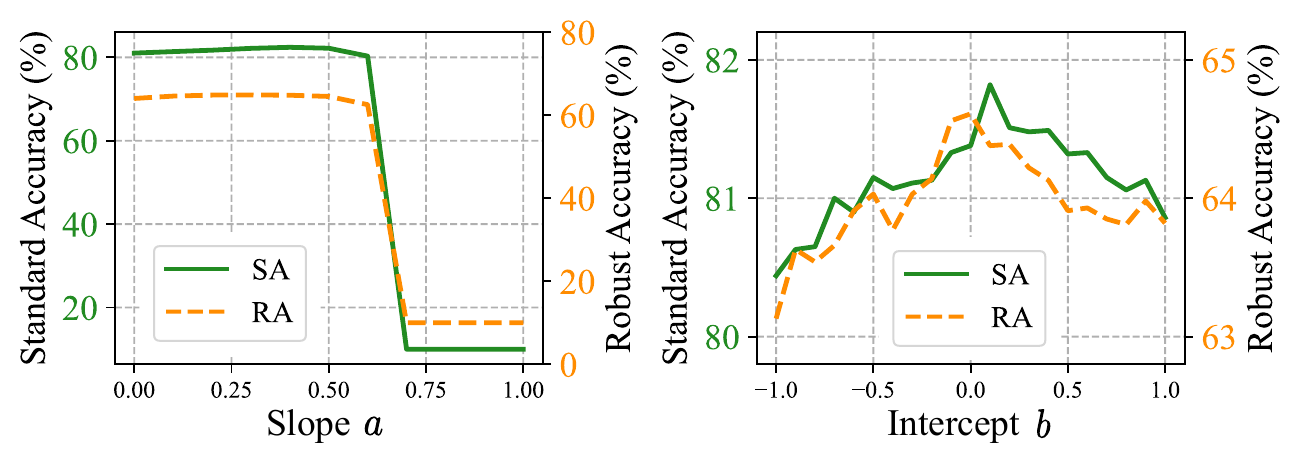}
\vspace{-9mm}
\caption{SA ($\%$) and RA ($\%$) over initialization of slope $a$ (\textit{Left}) and intercept $b$ (\textit{Right}) in grafting. }
\vspace{-3mm}
\label{fig:ab_init}
\end{figure}

\vspace{-2mm}
\paragraph{Initialization of grafting.} Experiments in Figure~\ref{fig:ab_init} demonstrate the effects of initialization for grafting's slope $a$ and intercept $b$. We found that the grafting performance is more sensitive to its slope especially for $a\in [0.5,0.8]$; $a\in[0.0,0.5]$, $b\in[-0.5,0.5]$ is the safe zone to initialize the grafted linear activation function. 

\paragraph{Whether fine-tuning weights in grafting?} After replacing ReLU with linear functions, there are two options of the fine-tuning: optimizing \{$a$, $b$\} only, or optimizing all \{$a$, $b$, $\mathbf{W}$\}. We conduct an ablation study in Table~\ref{tab:weight}, and observe that updating weights together achieves consistently better performance. It is within expectation since FAT fine-tuned weights preserve more empirical robustness (RA), leaving more room for VA improvements.

\begin{table}[!ht]
\centering
\vspace{-3mm}
\caption{Ablation on whether fine-tuning weights in grafting. Unstable neuron ratio (UNR $\%$), \colorbox{LightCyan}{VA} ($\%$), SA ($\%$), and RA ($\%$) of ConvBig with $50\%$ grafted neurons on CIFAR-10 are reported.}
\resizebox{\linewidth}{!}{
\begin{tabular}{@{}l|cacc}
\toprule
\multirow{1}[1]{*}{Settings} & UNR & VA & SA & RA \\ \midrule
Grafting (50$\%$) w. Tuning Weight  & 4.32 & 39.12 & 62.23 & 47.73 \\
Grafting (50$\%$) w.o. Tuning Weight & 5.24 & 36.20 & 57.60 & 44.33 \\
\bottomrule
\end{tabular}}
\label{tab:weight}%
\vspace{-3mm}
\end{table}%

\begin{figure}[!ht]
\centering
\vspace{-2mm}
\includegraphics[width=1\linewidth]{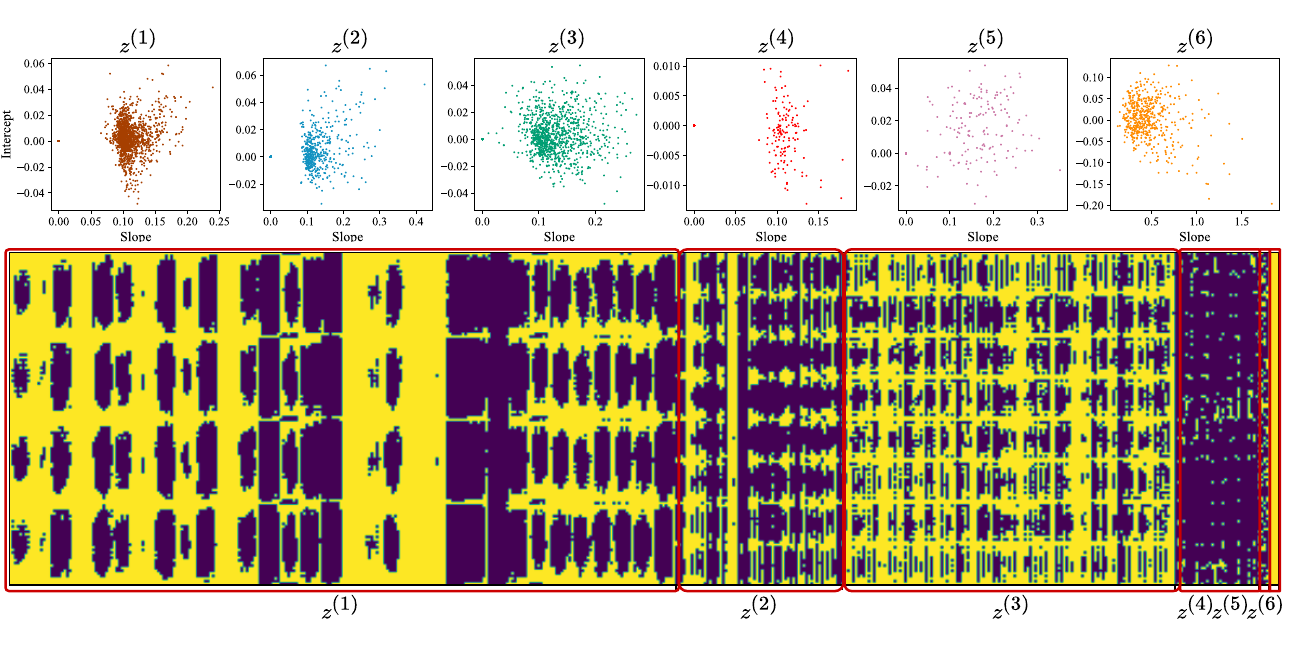}
\vspace{-11mm}
\caption{The \textit{Bottom} figure visualizes grafting masks from ConvBig with $50\%$ grafted neurons on CIFAR-10. The bright dots (\textcolor{yellow}{$\bullet$}) are the grafted neurons and the dark dots (\textcolor[RGB]{202,12,22}{$\bullet$}) stand for vanilla ReLU neurons. The \textit{Upper} figure shows the corresponding learned intercept over slope for each layer's activation, $z^{(1)}\sim z^{(6)}$. Please zoom-in for better readability.}
\vspace{-3mm}
\label{fig:mask}
\end{figure}

\paragraph{Visualization of grafting mask and its slope \& intercept.} As shown in Figure~\ref{fig:mask}, we visualize the obtained grafting mask and learned slopes and intercepts. The grafted neurons mainly scatter in the first few (i.e., $z^{(1)},z^{(2)},z^{(3)}$) and the last (i.e., $z^{(6)}$) layers, which appears to have block-wise patterns especially in $z^{(1)}$. The distributions of optimized slope and intercept are centered at $[0.1,0.3]$ and $0$, respectively.

\subsection{Additional Studies on Grafting} \label{sec:extra_study}

\paragraph{Combining with existing regularization.} Besides manipulating relaxed neurons, there exist several other effective regularizations~\citep{xiao2018training} for improving certifiable robustness: ($i$) ReLU stability regularizer that enforces ReLU to be always active or inactive given specific input perturbations; ($ii$) $\ell_1$ weight regularization; ($iii$) small weight pruning. Since they work orthogonally to our proposal, we show results of their combination in Table~\ref{tab:other_reg}. It suggests that grafting functions complementarily, and is capable of being integrated with previous techniques for further enhanced certifiable robustness.

\begin{table}[!ht]
\centering
\vspace{-3mm}
\caption{Combining grafting with existing regularizations for achieving better verified accuracy. Unstable neuron ratio (UNR $\%$), \colorbox{LightCyan}{VA} ($\%$), SA ($\%$), and RA ($\%$) of CNN-B with $50\%$ grafted neurons on CIFAR-10 are reported.}
\resizebox{\linewidth}{!}{
\begin{tabular}{@{}l|cacc}
\toprule
\multirow{1}[1]{*}{Settings} & UNR & VA & SA & RA \\ \midrule
Grafting (50$\%$) & 5.36 & 50.40 & 74.08 & 58.76 \\
Grafting (50$\%$) + ReLU Stability Reg. & 5.71 & 51.90 & 75.95 & 60.77 \\
Grafting (50$\%$) + $\ell_1$ Reg. & 5.83 & 51.90 & 76.31 & 61.10  \\
Grafting (50$\%$) + Small Weight Pruning & 5.88 & 52.50 & 76.29 & 60.85 \\ \midrule
Grafting (50$\%$) + All & 5.73 & 52.55 & 76.15 & 61.08  \\
\bottomrule
\end{tabular}}
\vspace{-4mm}
\label{tab:other_reg}%
\end{table}%

\vspace{-1mm}
\paragraph{Different methods to locate insignificant neurons.} To determine the unnecessary portion of neurons and compute scores $r_s$, we can either choose some heuristics like activation magnitude (SAP)~\citep{dhillon2018stochastic} and gradient (GAP)~\citep{ye2020accelerating}, or learned importance scores (Hydra)~\citep{sehwag2020hydra}. A comprehensive comparison of these activation pruning algorithms has been carried out in Figure~\ref{fig:pruning}, where random pruning and dense networks as two baselines are included. We find that GAP consistently surpasses other approaches at a large range of activation sparsity from $0\%$ to $85\%$, while the learning-based sparsification shows moderate advantages at the extreme sparsity. Therefore, our grafting pipeline adopts the best-performing ``activation gradient'' (GAP)~\citep{ye2020accelerating} as a proxy for the neuron's significance. 

\vspace{-2mm}
\paragraph{Grafting $0$ versus $a\times x+b$.} Pruning (zeroing out) neurons is a spacial case of our grafting, where $\mathcal{A}_g(x):=0$. From Table~\ref{tab:pruning}, we see grafting learnable linear functions obtains superior performance in terms of all evaluation metrics. Note that unlike classical activation pruning methods which only depend on significance scores $r_s$ (Table~\ref{tab:main_res_2}), in the case of grafting zero, we still consider both significance and instability scores for picking candidate neurons in the same way when we graft $a\times x+b$, e.g., $\gamma\times r_u-r_s$.

\begin{table}[!ht]
\centering
\vspace{-2mm}
\caption{Comparison to grafting zero and gradually grafting. UNR ($\%$), \colorbox{LightCyan}{VA} ($\%$), SA ($\%$), and RA ($\%$) of CNN-B with $50\%$ grafted neurons on CIFAR-10 are reported.}
\resizebox{1\linewidth}{!}{
\begin{tabular}{@{}l|cacc}
\toprule
\multirow{1}[1]{*}{Settings} & UNR & VA & SA & RA \\ \midrule
Baseline & 15.85 & 37.40 & 79.95 & 62.23 \\ \midrule
Grafting ($\mathcal{A}_g(x)=0$) & 5.60 & 48.40 & 73.19 & 57.78 \\
Grafting ($\mathcal{A}_g(x)=a\times x+b$) & 5.36 & 50.40 & 74.08 & 58.76 \\
Gradually Grafting ($\mathcal{A}_g(x)=a\times x+b$) & 5.91 & 49.00 & 75.43 & 58.79 \\ 
\bottomrule
\end{tabular}}
\vspace{-4mm}
\label{tab:pruning}%
\end{table}%

\vspace{-1mm}
\paragraph{One-shot versus gradually grafting.} Although our main results are produced by one-shot grafting, it can be easily extended into a gradual style by assigning the desired amount of linearity to several iterations. Specifically, we adopt the schedule from~\citet{zhu2017prune} and graft a small portion of neurons in each iteration of the first half training epochs (i.e., $100$ epochs). Table~\ref{tab:pruning} shows that gradual grafting tends to maintain more standard accuracy by using more unstable ReLU neurons while sacrificing some verified accuracy. 

\vspace{-2mm}
\paragraph{Comparison with classical certified robust training.} One of our goals is to achieve satisfying certifiable robustness and avoid time-consuming certified robust training. Supportive results demonstrated in Table~\ref{tab:certified_training} again validate our proposed grafting pipeline. It shows that \textit{FAT + Grafting} obtains $2.85\%$ VA, $16.08\%$ SA, $10.14\%$ RA improvements with $92.87\%$ training time savings, compared to the Auto-LiRPA based certified robust training~\citep{xu2020automatic} with a target perturbation radius $\epsilon=2/255$. 

\begin{table}[!ht]
\centering
\vspace{-2mm}
\vspace{-2mm}
\caption{Comparisons between a representative certified robust training method~\citep{xu2020automatic}, and our grafting with fast adversarial training (FAT) without explicit certified defense training. Here we report UNR ($\%$), \colorbox{LightCyan}{VA} ($\%$), SA ($\%$), RA ($\%$), and total training time (hour) of CNN-B w./w.o. $50\%$ grafted neurons on CIFAR-10 are reported.}
\resizebox{1\linewidth}{!}{
\begin{tabular}{@{}l|caccc}
\toprule
\multirow{1}[1]{*}{Settings} & UNR & VA & SA & RA & Training Time \\ \midrule
Baseline (FAT) & 15.85 & 37.40 & 79.95 & 62.23 & 0.39 h \\ \midrule
Certified Robust Training & 0.96 & 47.55 & 58.00 & 48.62 & 16.26 h \\ 
FAT + Grafting (50\%) & 5.36 & 50.40 & 74.08 & 58.76 & 1.13 h  \\
\bottomrule
\end{tabular}}
\vspace{-2mm}
\label{tab:certified_training}%
\end{table}%

\section{Conclusion}
\vspace{-1mm}
This paper proposes \textit{linearity grafting}, a new relaxed neuron pruning paradigm for improved certifiable robustness and verification scalability.
Specifically, it grafts appropriate forms of linearity by replacing the redundant non-linear activation functions. The benefits come from: ($1$) grafting insignificant and unstable neurons directly reduces the harmful and excessive non-linearity and tightens the verification bounds; ($2$) the further optimized parameters (slopes and intercepts) in grafted linear neurons help maintain good generalization performance and avoid ill-conditioned activation states (e.g., most ReLU neurons are inactive and fixed at 0). For future works, grafting also motivates us to design verification-friendly architectures with a mixture of linear and non-linear activation functions.

\vspace{-0.5em}
\section*{Acknowledgment}
\vspace{-0.5em}
{\small Z.W. is in part supported by an NSF SCALE MoDL project (\#2133861).}


\bibliography{Grafting}
\bibliographystyle{icml2022}

\clearpage

\newpage
\appendix
\renewcommand{\thepage}{A\arabic{page}}  
\renewcommand{\thesection}{A\arabic{section}}   
\renewcommand{\thetable}{A\arabic{table}}   
\renewcommand{\thefigure}{A\arabic{figure}}


\section{More Experiment Results} \label{sec:more_res}

\paragraph{Different target perturbation radius $\epsilon$.} We validate the effectiveness of our proposed grafting across different target perturbation radius $\epsilon=\frac{8}{255}$, which is usually a more challenging setup. Experiment results are collected in Table~\ref{tab:main_res_8}, which consistently demonstrate the superiority of grafted NN in terms of boosted VA and reduced UNR. 

\begin{table}[!ht]
\centering
\vspace{-2mm}
\caption{\textbf{Unstable neuron ratio (UNR $\%$), verified accuracy (\colorbox{LightCyan}{VA} $\%$), standard accuracy (SA $\%$), PGD-$100$ robust accuracy (RA $\%$), and average time (s)} of FAT trained models w./w.o. grafting on (CIFAR-10, CNN-B). $\beta$-CROWN~\citep{wang2021beta} with BaB, the current SOTA complete verifier is used to compute VA. \textbf{The target perturbation size is $\epsilon=\frac{8}{255}$.}}
\resizebox{\linewidth}{!}{
\begin{tabular}{@{}l|caccc}
\toprule
\multicolumn{1}{c|}{\multirow{2}[2]{*}{FAT ($\epsilon=\frac{8}{255}$)}} & \multicolumn{5}{c}{(CNN-B, CIFAR-10)}  \\ \cmidrule{2-6}
\multicolumn{1}{c|}{} & UNR & VA & SA & RA & Time  \\ \midrule
Baseline & 37.44 & 0.30 & 66.47 & 34.23 & 300.88 \\ \midrule
SAP~\citep{dhillon2018stochastic} ($50\%$) & 18.20 & 0.90 & 63.87 & 32.66 & 295.33 \\
GAP$^\dagger$~\citep{ye2020accelerating} ($50\%$) & 23.83 & 0.40 & 65.91 & 33.46 & 300.46 \\
Hydra$^\ddagger$~\citep{sehwag2020hydra} ($50\%$) & 20.12 & 0.40 & 64.38 & 31.49 & 283.44 \\ \midrule
Random Grafting ($50\%$) & 20.12 & 2.00 & 60.41 & 31.66 & 283.44 \\ 
Grafting ($50\%$) & 12.35 & 4.70 & 58.87 & 31.34 & 257.59 \\ \midrule
Grafting ($30\%$) & 17.47 & 1.10 & 64.46 & 32.83 & 294.55 \\
Grafting ($80\%$) & 4.93 & 14.90 & 46.76 & 27.25 & 112.11 \\
\bottomrule
\end{tabular}}
\label{tab:main_res_8}%
\begin{tablenotes}
\footnotesize
\item $\dagger$ The heuristic of activation gradient magnitude~\citep{ye2020accelerating} is utilized to guide activation pruning. 
\item $\ddagger$ Based on the official implementation of~\citet{sehwag2020hydra}, we extend the original sparse mask learning to activation. 
\end{tablenotes}
\vspace{-4mm}
\end{table}%

\begin{figure}[!ht]
\centering
\vspace{-2mm}
\includegraphics[width=1\linewidth]{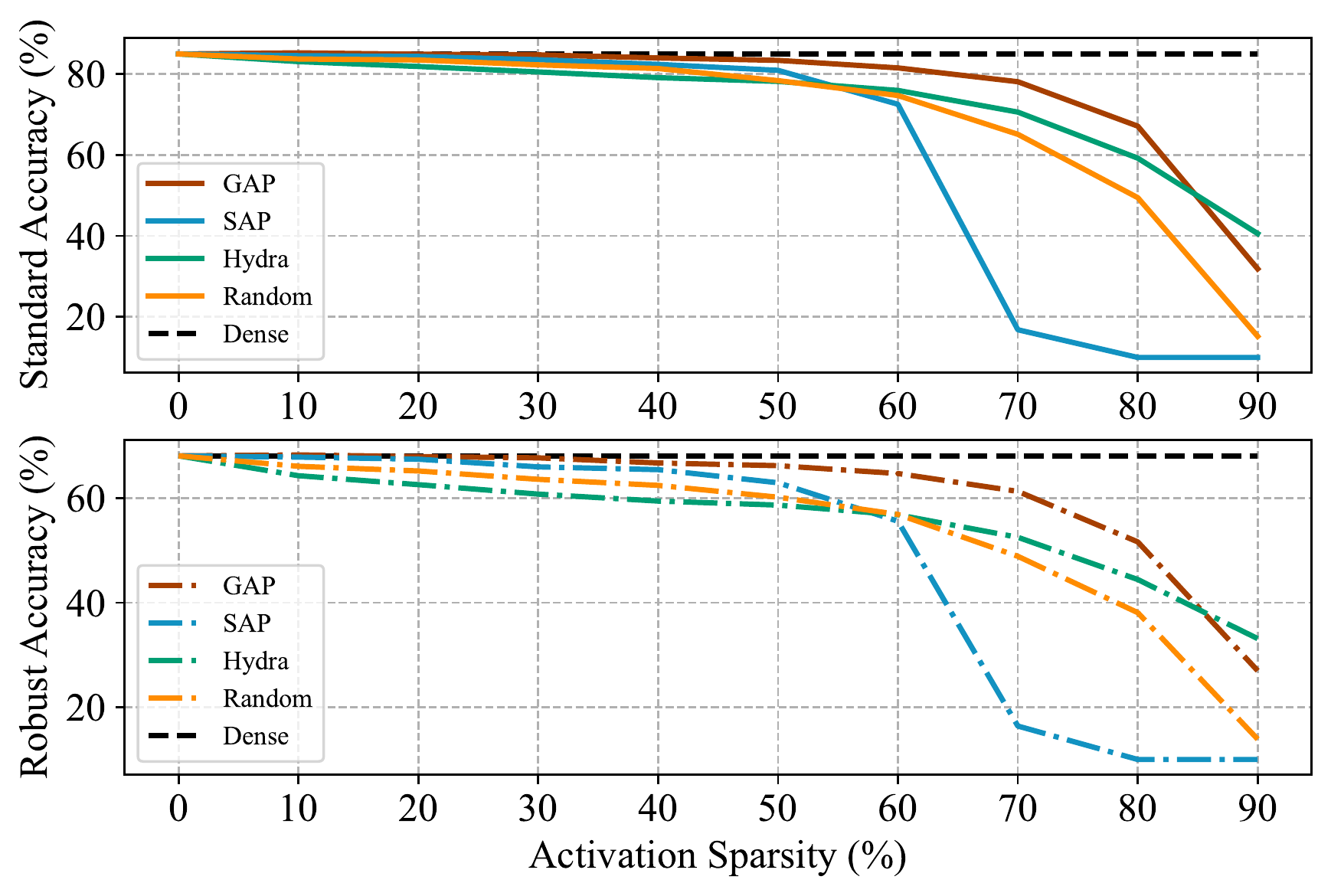}
\vspace{-8mm}
\caption{Performance (SA $\%$ and RA $\%$) of activation pruning with diverse heuristics or optimized scores, which lies the foundation of locating insignificant neurons.}
\vspace{-4mm}
\label{fig:pruning}
\end{figure}

\paragraph{Comparison of different neuron significance proxies.} As shown in Figure~\ref{fig:pruning}, the activate gradient of GAP does the best job of approximating the neuron significance.

\paragraph{Corner test cases.} We conduct corner case evaluations for our grafted NNs, where pruning identified exemplars (PIE)~\cite{hooker2019compressed} are adopted as the corner test cases. From Table~\ref{tab:corner}, we observe grafting still obtains substantial VA boosts ($+11.71\%$) on corner cases with a moderate decrease in SA.
\begin{table}[!ht]
\centering
\vspace{-2.5mm}
\caption{Corner case study. \textbf{Verified accuracy (VA $\%$) and standard accuracy (SA $\%$)} of FAT trained models w./w.o. grafting on (CIFAR-10, CNN-B) are reported.}
\label{tab:corner}
\resizebox{1\linewidth}{!}{
\begin{tabular}{@{}l|cccc}
\toprule
\multirow{1}{*}{(C10, CNN-B)} & \multirow{1}{*}{VA (corner cases)} & \multirow{1}{*}{SA (corner cases)} & \multirow{1}{*}{VA (regular testset)} & \multirow{1}{*}{SA (regular testset)}\\  
\midrule
Baseline & $9.01\%$ & $54.05\%$ & $37.40\%$ & $79.95\%$  \\
Grafting ($50\%$) & $20.72\%$ ($\textcolor{red}{+11.71\%}$) & $51.35\%$ ($\textcolor{blue}{-2.70\%}$) & $50.40\%$ ($\textcolor{red}{+13.00\%}$) & $74.08\%$ ($\textcolor{blue}{-5.87\%}$) \\
\bottomrule
\end{tabular}}
\vspace{-4mm}
\end{table}%

\paragraph{Investigations on non-ReLU networks.} Our idea of replacing highly non-linear activation functions with linear operations to ease verification is general, and can be applied to other non-linear functions as long as the verifier supports them. We demonstrate \emph{preliminary results} on a CNN using \textbf{sigmoid} activation function, with improved verified accuracy (VA) after grafting shown in Table~\ref{tab:sigmod}. It is less improved compared to ReLU networks mostly because current verifiers have limited support for non-ReLU activation.
\begin{table}[!ht]
\centering
\vspace{-7mm}
\caption{Studies of non-ReLU networks. \textbf{Verified accuracy (VA $\%$)} of FAT trained models w./w.o. grafting on (CIFAR-10, Sigmod 3-layer CNN) are reported.}
\label{tab:sigmod}
\resizebox{1\linewidth}{!}{
\begin{tabular}{@{}l|ccc}
\toprule
\multirow{1}{*}{(CIFAR-10, Sigmoid 3-layer CNN)} & Baseline & \textbf{Grafting} ($50\%$) & GAP ($50\%$) \\  
\midrule
 Verified Accuracy (VA) & $38.62\%$ & $39.46\%$ ($\textcolor{red}{+0.84\%}$) & $35.41\%$ ($\textcolor{blue}{-3.21\%}$) \\
\bottomrule
\end{tabular}}
\vspace{-6mm}
\end{table}%

\paragraph{Linearity accelerates the verification?} Our grafted linearity greatly reduced the relaxation and branching needed in verifiers, decreasing the verification complexity exponentially, thus more examples can be verified within a short time. We do not accelerate model computation, and in fact we use an extra binary mask to indicate the grafted neurons which slightly increases computation during training (we measured $0.076$s $\to$ $0.123$s per batch for ConvBig). 

\begin{figure}[!ht]
\centering
\vspace{-2mm}
\includegraphics[width=0.65\linewidth]{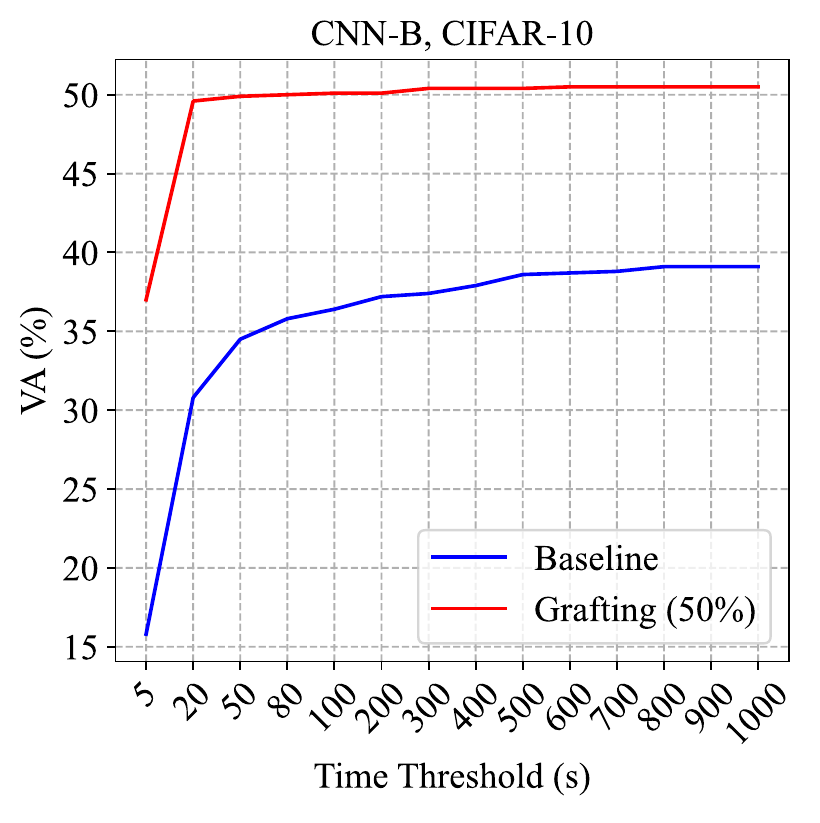}
\vspace{-5mm}
\caption{\textbf{VA $\%$} under different maximum time thresholds.}
\vspace{-4mm}
\label{fig:time}
\end{figure}

\paragraph{Results with the increased verification time.} In Figure~\ref{fig:time}, we plot the number of verified examples with $1,000$s verification time. Since grafting reduces verification complexity exponentially, a verifier can quickly verify much more examples than the baseline within a short time.

\paragraph{Societal impact.} We believe our work improves the scalability of certification and therefore has overall positive societal impacts on various safety-crucial applications. However, it may potentially be misused to identify the weakness of DNNs and guide malicious attacks.

\end{document}